\documentclass[runningheads]{llncs}

\usepackage{booktabs}
\usepackage{amsmath}
\usepackage{amssymb}
\usepackage{hyperref}
\usepackage{url}
\usepackage{subfigure}
\usepackage{graphicx}
\usepackage{tikz}
\usetikzlibrary{shapes.geometric, arrows}
\usepackage{color}

\usepackage[misc]{ifsym}
\newlength\lengtha 
\newlength\lengthb 
\newcommand{\tabincell}[2]{\begin{tabular}{@{}#1@{}}#2\end{tabular}}
\newcommand{\tablefont}{\fontsize{8pt}{\baselineskip}\selectfont}

\begin{document}
\title{Deep Detection for Face Manipulation}
\author{Disheng Feng\inst{1}
\and
Xuequan Lu\inst{1}\orcidID{0000-0003-0959-408X}{\Letter}
\and
Xufeng Lin\inst{1}
}
\authorrunning{D. Feng et al.}

\institute{Deakin University, Australia\\
\email{sheldonvon@outlook.com}\\
\email{\{xuequan.lu,xufeng.lin\}@deakin.edu.au}
}
\maketitle              
\begin{abstract}
It has become increasingly challenging to distinguish real faces from their visually realistic fake counterparts, due to the great advances of deep learning based face manipulation techniques in recent years. In this paper, we introduce a deep learning method to detect face manipulation. It consists of two stages: feature extraction and binary classification. To better distinguish fake faces from real faces, we resort to the triplet loss function in the first stage. We then design a simple linear classification network to bridge the learned contrastive features with the real/fake faces. Experimental results on public benchmark datasets demonstrate the effectiveness of this method, and show that it generates better performance than state-of-the-art techniques in most cases.
\end{abstract}
\section{Introduction}
\label{sec:introduction}
Face manipulation has received considerable attention in recent years. Its detection or localization becomes more challenging with the rapid development of deepfake generation techniques. For example, Karras et al. \cite{karras2018progressive} introduced a high-resolution full-forged image synthesis method by progressively adding layers to both the generator and discriminator. 
Thies et al. \cite{thies2019deferred} introduced a deferred neural rendering technique that equips the traditional graphics pipeline with learnable components for synthesizing photo-realistic face images. The increasing appearance of such computer-generated face images or videos has cast serious doubts on the credibility of digital content, and potentially poses huge threats to privacy and security. Thus, it is essential to devise robust and accurate detection methods to identify face manipulations.

Some existing methods detect face manipulation based on the visual cues left by the manipulation methods, such as the abnormality of head pose \cite{Yang2019ExposingDF}, the frequency of eye blinking \cite{Blinking2018}, and the face warping artifacts \cite{Li2018ExposingDV}. However, these methods are designed for specific visual disturbance caused by manipulation methods, which may fail as the manipulation technologies become more mature in the foreseeable future. Other detection approaches resort to deep learning techniques \cite{afchar2018mesonet,dang2019detection,nguyen2019use,rssler2019faceforensics,tolosana2020deepfakes}. 
Nevertheless, these methods usually ignore the role of loss functions, which consequently hinders the improvement of detection performance. 
Furthermore, most deep learning models for face manipulation detection are trained and tested on the same dataset, which overlooks the cross-dataset adaption of the trained model and brings difficulties for other researchers to perform reasonable comparisons.

Given the above motivations, we introduce a novel face manipulation detection approach in this paper. Our key idea is to use a loss function for contrastive learning of three samples. We achieve this by resorting to the triplet loss \cite{hoffer2014deep}. Specifically, inspired by \cite{hoffer2014deep}, we design three samples with two positive samples (one is anchor) and one negative sample, and then formulate the loss by minimizing the distance between one positive sample and the anchor and maximizing the distance between the anchor and the negative sample. We employ Xception \cite{chollet2016xception} as the backbone, with certain modifications (Section. \ref{sec:networkstructure}). We further design a linear classification network to perform binary classification (0 and 1, corresponding to real and fake faces respectively), to bridge the learned contrastive features with the real or fake faces. In addition to the intra-dataset evaluation, we also conduct the cross-dataset evaluation (Section \ref{subsec:Cross-dataset}). Experiments validate our method and demonstrate that it outperforms state-of-the-art techniques in most cases, in terms of detection performance evaluated with AUC (Area Under the ROC Curve). We also conduct an ablation study showing that our contrastive learning enables substantial improvement in detection accuracy.

\section{Related Work}
\label{sec:relatedwork}
In this section, we will review the works that are most related to our work. We first revisit the previous research on face manipulation methods and then cover the current feasible detection methods.

\textbf{Face image manipulation methods.} 
The origin core mind of deepfake can be found in 2016, when Zhmoginov and Sandler \cite{zhmoginov2016inverting} developed a gradient-ascent approach to effectively invert low-dimensional face embeddings to realistically looking consistent images, which is also based on GAN. This technique was later implemented in the well-known mobile phone application called FaceAPP \cite{Faceapp2020} which contributes to some of the face forgery datasets. Thies et al. \cite{thies2019deferred} introduced a deferred neural rendering method called Neural Texture which can synthesize face manipulate images in conjunction with rendering network. Li et al. \cite{Celeb_DF_cvpr20} proposed an enhanced DeepFake-based method to solve problems such as visible splicing boundaries, color mismatch and visible parts of original face, which was used to generate the Celeb-DF dataset.

\textbf{DeepFake Detection Methods.} In recent years, some methods have been developed to detect the visual imperfections left by Deepfake algorithms. Li et al. \cite{Li2018ExposingDV}  presented a detection method against the face-warping drawback of the Deepfake methods (caused by limited synthesis resolution). They proposed another method \cite{Blinking2018} to detect the eye blinking details which is a physiological signal that could be easily neglected by Deepfake methods. Yang et al. \cite{Yang2019ExposingDF} proposed a detection approach using the inconsistencies in the head poses of the fake videos. Matern et al. \cite{Exploiting2019} proposed a similar detection method regarding a variety of visual features. However, one obvious drawback is that those visual imperfections could be easily refined \cite{Celeb_DF_cvpr20}, and thus they could be invalid very soon. On the other hand, some deep learning methods are proposed to deal with the Deepfake detection problem. Zhou et al. \cite{zhou2018twostream} presented a method using two-stream GoogLeNet InceptionV3 model \cite{szegedy2014going} to achieve state-of-the-art performance. Afchar et al. \cite{afchar2018mesonet} offered a feasible method called MesoNet, providing an important baseline for comparisons. Rössler et al. \cite{rssler2019faceforensics} presented high performance results using network structure based on depth-wise separable convolution layers with residual connections called Xception \cite{chollet2016xception}. Some more recent methods such as Dang et al. \cite{dang2019detection} and Tolosana et al. \cite{tolosana2020deepfakes} also utilized Xception as the backbone network. Nguyen et al. \cite{nguyen2019use} used a capsule structure with a VGG19 as an embedded backbone network for DeepFake detection. Nguyen et al. \cite{nguyen2019multitask} proposed to use multi-task learning approach to simultaneously detect manipulated data and locate the manipulated regions for each query.  

\section{Our Method}
\label{sec:method}
Our method first learns features with the aid of a contrastive loss (i.e., triplet loss) which can metrically push the forgery faces away from the real faces. Then we design a simple linear classification network to obtain binary results. 

\subsection{Network Structure}
\label{sec:networkstructure}
\textbf{Feature extraction network.}
We first introduce the feature extraction stage. We need to train a function $f(\theta):\mathbb{R}^D \to \mathbb{R}^E$, mapping semantically different data $\mathbb{R}^D$ to metrically further points $\mathbb{R}^E$. Following the work \cite{hoffer2014deep}, we construct a model satisfying 
\begin{equation}\label{eq:tripletequation}
\begin{aligned}
TripletNet(x, x^-, x^+) = 
\begin{bmatrix}
\|Net(x)-Net(x^-)\|_2
\\
\|Net(x)-Net(x^+)\|_2
\end{bmatrix}
\in \mathbb{R}^2_+,
\end{aligned}
\end{equation}
where $Net(x) = Xception(x)$ and $x$ is a randomly selected image (the anchor, could be manipulated image or pristine image). $x^+$ and $x^-$ are images which are randomly selected from the same class as the anchor and a different class, respectively. We simply employ the Xception \cite{chollet2016xception} as the backbone network, since it has been proved to be effective and powerful. To achieve this, we modify the last layer with a dropout layer in conjunction with a linear layer.

\textbf{Classification network.}
We devise a simple linear classification network that takes the output of the feature extraction network (2D points) as input and outputs a binary value (0 or 1) to indicate real or fake. The components of this network are as follows: Linear layer (size:2), ReLu layer, Linear layer (size:128), Linear layer (size:256), ReLu layer, Linear layer (size:128), ReLu layer, Linear layer (size:2), Leaky ReLu layer.

\subsection{Training and Inference}
We first train the feature extraction network and use the well-tuned backbone network to extract the features from the images. We then train the linear classification network using those extracted features (2D points) as train data. Notice that this process is also used for the inference stage.

\subsection{Implementation}
Our networks are implemented in PyTorch on a desktop PC with an Intel Core i9-9820X CPU (3.30GHz, 48GB memory) and a GeForce RTX 2080Ti GPU (11GB memory, CUDA 10.0). The SGD optimizer is used for training both the feature extraction and classification networks. Specifically, the former is trained with a learning rate of $4\times 10^{-4}$ and the batch size is typically set to $12$ in our experiments. We observed that the loss became steady after $8$ epochs. Therefore, the number of epochs is empirically set to $10$. The classification network is trained with a learning rate and momentum of $3.0\times 10^{-3}$ and $1.0\times 10^{-1}$, respectively. 

\section{Experimental Results}
\label{sec:Results}
In this section, we will first introduce the datasets used in our experiments. We then report the results and analyze the effectiveness of our model by evaluating the triplet structure. In the end, we will compare our model with state-of-the-art methods in both intra-dataset setting and cross-dataset setting.

\subsection{Datasets}
\label{subsec:datasets}
A variety of face manipulation datasets have been proposed to facilitate the evaluation and benchmarking of image manipulation detection methods. In our experiments, we used the following datasets: \textit{FaceForensics++ (FF++)} \cite{rssler2019faceforensics}, \textit{UADFV} \cite{Li2018ExposingDV} and \textit{Celeb-DF} \cite{Celeb_DF_cvpr20}. Note that, \textit{FF++} includes DeepFake videos  generated with four face manipulation methods (Deepfake \cite{Deepfakes2020}, Neural Texture \cite{thies2019deferred}, faceswap \cite{Faceswap2020} and face2face \cite{Face2Face2016}). \textit{Celeb-DF} is a large face forgery dataset with refined video synthesis solving common problems such as temporal flickering frames and color inconsistency in other datasets.

Similarly, the above datasets provide data in the format of videos. To train our network, we use the face detector in Dlib \cite{dlib09} to crop the video frames into a specific size. Detail information can be found in Table \ref{table:amountdetail}.

\begin{table}[thbp]\tablefont
    \begin{center}
    \caption{Image numbers in the split sets used in our experiments.}\label{table:amountdetail}
    \begin{tabular}{@{} l
                @{\hspace*{\lengtha}}c
                @{\hspace*{\lengtha}}c
                @{\hspace*{\lengtha}}c
                @{\hspace*{\lengtha}}c}
    \toprule
    Datasets & \tabincell{l}{
    Train (real)} & \tabincell{l}{
    Train (fake)
    } & \tabincell{l}{
    Test (real)
    } & \tabincell{l}{Test (fake)}\\ 
    \midrule
    FF++ & 115556 & 108935 & 20393 & 20473
    \\
    UADFV & 10100 & 9761 & 1783 & 1723
    \\
    Celeb-DF & 172187 & 165884 & 30386 & 29259
    \\
    \bottomrule
    \end{tabular}
    \end{center}
\end{table}

\subsection{Extracted Features}
\label{sec:features}
In this section, we present the visual results of the extracted features. We only report the results on the FF++ dataset for illustration purpose. We generate one dataset for each method (Neural Texture, DeepFake, Face2Face and DeepFake), including real faces and corresponding manipulated faces. 

\begin{figure}[hbt!]
\centering
\begin{minipage}[b]{0.15\linewidth}
\subfigure[]{\label{}\includegraphics[width=1\linewidth]{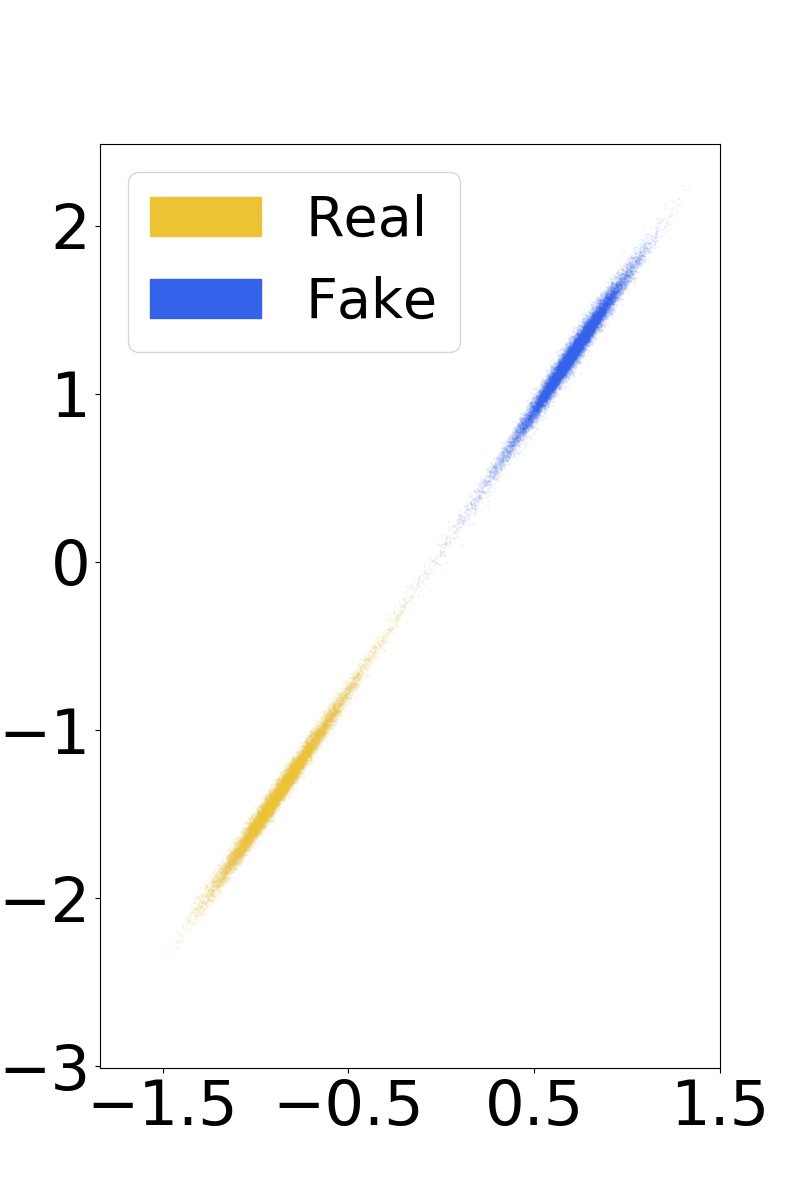}}
\end{minipage}
\begin{minipage}[b]{0.15\linewidth}
\subfigure[]{\label{}\includegraphics[width=1\linewidth]{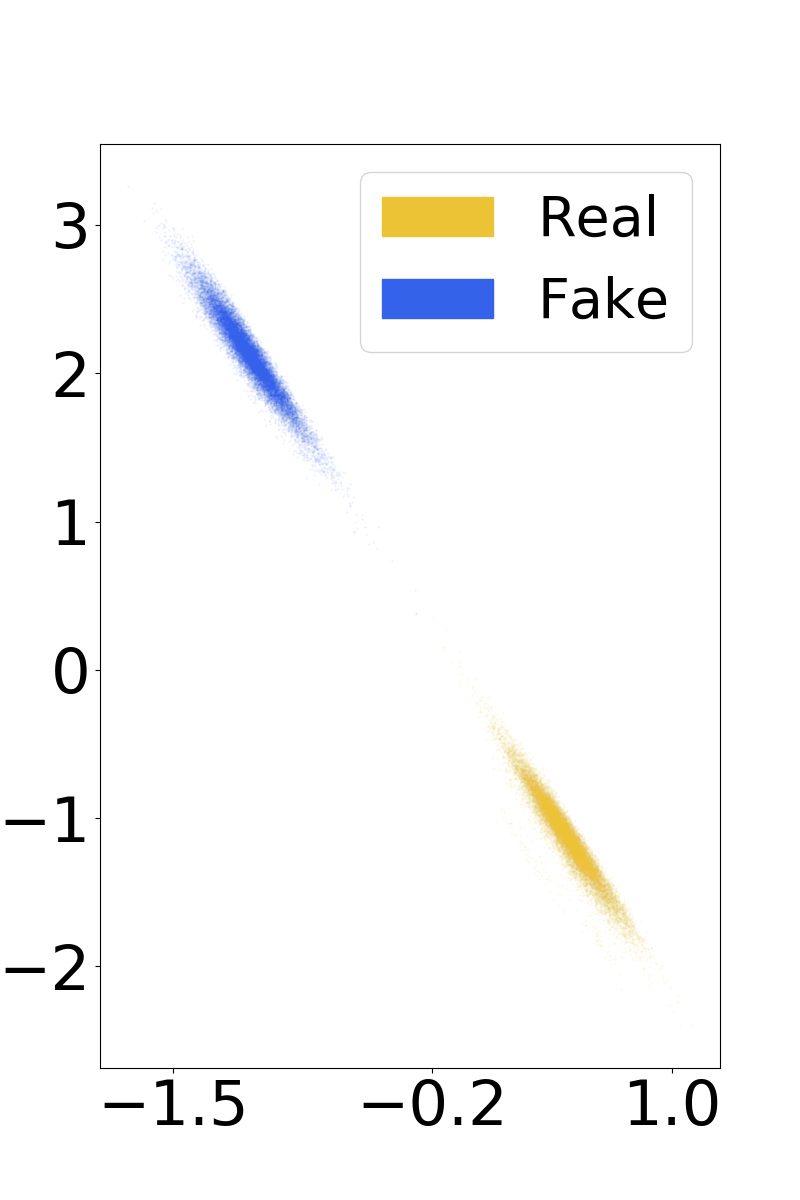}}
\end{minipage}
\begin{minipage}[b]{0.15\linewidth}
\subfigure[]{\label{}\includegraphics[width=1\linewidth]{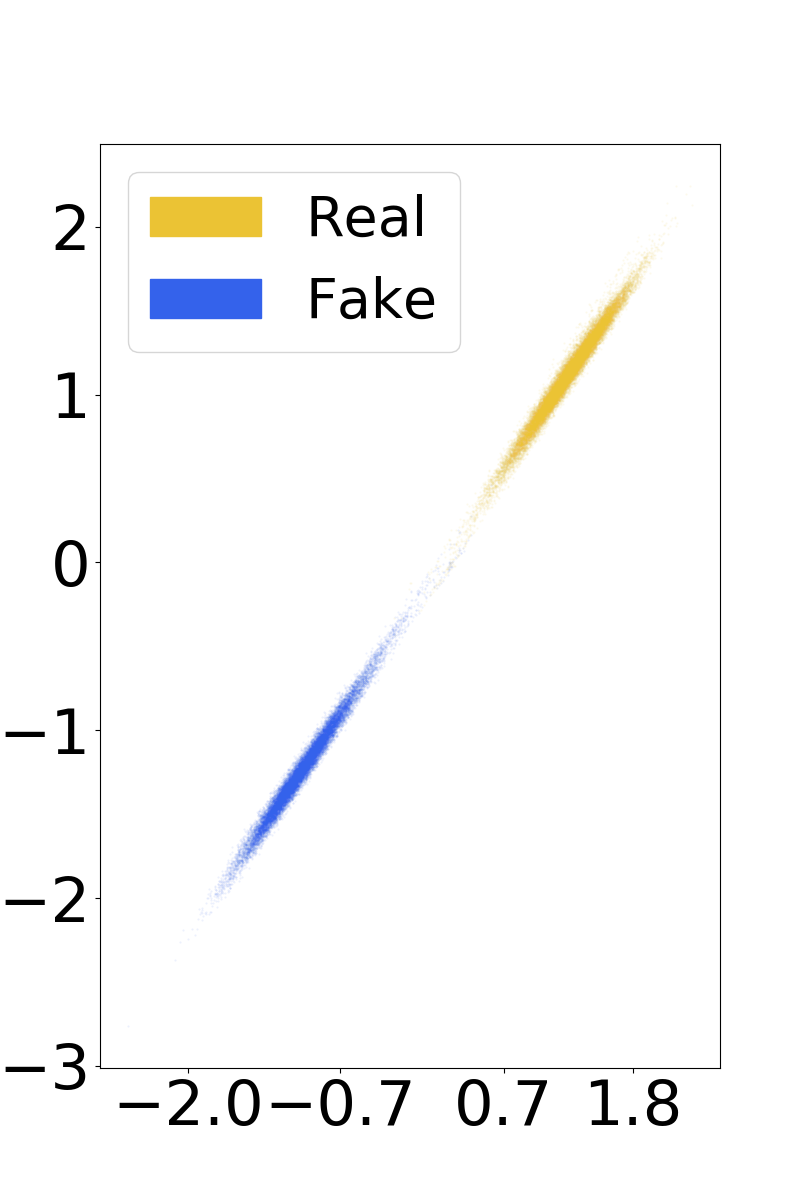}}
\end{minipage} 
\begin{minipage}[b]{0.15\linewidth}
\subfigure[]{\label{}\includegraphics[width=1\linewidth]{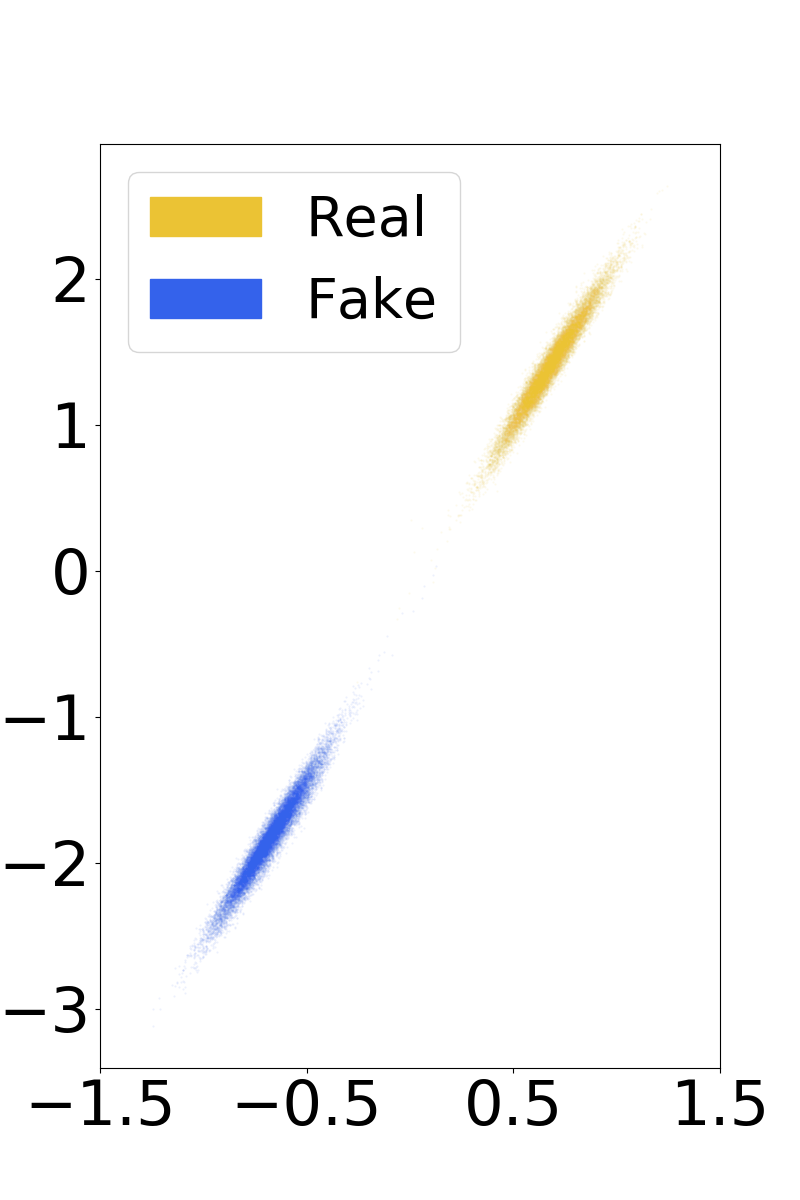}}
\end{minipage}
\begin{minipage}[b]{0.15\linewidth}
\subfigure[]{\label{}\includegraphics[width=1\linewidth]{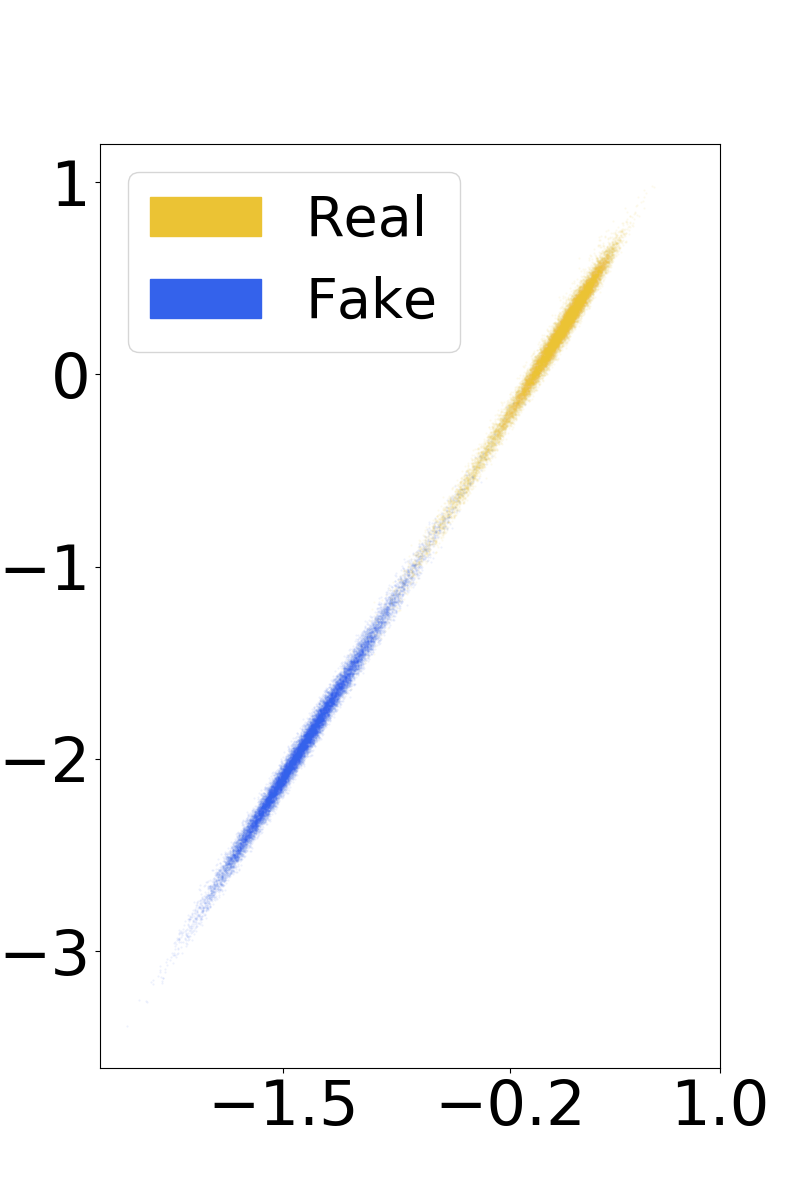}}
\end{minipage}
\caption{(a)-(e) are the outputs of the feature extraction network tested on five datasets (Neural Texture, DeepFake, Face2Face, FaceSwap and Combination, respectively).}
\label{fig:testEmbeddings}
\end{figure}

From  Fig. \ref{fig:testEmbeddings}, We observed that two different colors of points forming two distinct clusters. The results of the datasets using DeepFake and FaceSwap outperform the others.

In Table \ref{table:AUCandEER}, we list the Area Under the ROC Curve (AUC) as well as the Equal Error Rate (ERR) tested on the test sets of different datasets. We can see an overall near-perfect performance with extremely low EERs.

\begin{table}[hbt!]\tablefont
    \centering
    \caption{Comparisons of AUC (Area Under the ROC Curve) and EER (Equal Error Rate) on each dataset.  }\label{table:AUCandEER}
    \begin{tabular}{@{} l
                @{\hspace*{\lengtha}}c
                @{\hspace*{\lengtha}}c
                @{\hspace*{\lengtha}}c
                @{\hspace*{\lengtha}}c
                @{\hspace*{\lengtha}}c}
    \toprule
    Datasets & \tabincell{l}{
    DeepFake} & \tabincell{l}{
    Face Swap} & \tabincell{l}{
    Face2Face} & \tabincell{l}{
    Neural Texture} & \tabincell{l}{
    Combination}\\ 
    \midrule
    AUC & 99.99 & 99.99 & 99.99 & 99.99 & 99.96\\
    ERR & $9.47\times10^{-4}$ &  $9.72\times10^{-3}$ & $1.12\times10^{-1}$ &  $5.91\times10^{-2}$ &  $8.55\times10^{-1}$\\
    \bottomrule
    \end{tabular}
\end{table}
\subsection{Ablation Study}
We compare the Xception network and our proposed model (i.e., Xception network with the triplet loss), on four datasets generated in Section \ref{sec:features}. Although the Xception network itself is highly competent to detect the manipulated face images (98.85\%, 98.23\%, 98.36\%, 94.50\% for DeepFake, FaceSwap, Face2Face and NeuralTexture respectively), our contrastive learning with triplet loss clearly enables the learning of more discriminative features and boost the detection accuracy (1\%, 0.87\%, 1.34\% and 2.92\% higher than the corresponding methods respectively).
\subsection{Comparisons with State-of-the-art Methods}
\label{subsec:Cross-dataset}
In Table \ref{table:crossdatasets}, we show the comparisons of our method and the state-of-the-art methods on the above mentioned three datasets in both intra-dataset and cross-dataset settings. Specifically, we separately trained our networks on the train set of each dataset and evaluate the detection performance on the test sets of all three datasets. Although the results of our method and other methods are not strictly comparable, it provides reasonable comparisons \cite{dang2019detection}.

\begin{table}[hbt!]\tablefont
    \centering
    \caption{AUC(\%) on FF++, UADFV and Celeb-DF. Best results in intra-dataset setting are underlined, and best results for the cross-dataset setting are in bold. 
    }\label{table:crossdatasets}
    \begin{tabular}{@{} l
                @{\hspace*{\lengthb}}c
                @{\hspace*{\lengthb}}c
                @{\hspace*{\lengthb}}c
                @{\hspace*{\lengthb}}c
                @{\hspace*{\lengthb}}c}
    \toprule
    Methods & \tabincell{l}{
    Train data
    } & \tabincell{l}{
    FF++
    } & \tabincell{l}{
    UADFV
    } & \tabincell{l}{
    Celeb-DF
    }\\ 
    \midrule
    Two-stream\cite{zhou2018twostream} & Private & 70.1 & 85.1 & 53.8
    \\
    Meso4\cite{afchar2018mesonet} & Private & 84.7 & 84.3& 54.8
    \\
    MesoInception4\cite{afchar2018mesonet} & Private & 83.0 & 82.1 & 53.6
    \\
    HeadPose\cite{Yang2019ExposingDF} & UADFV & 47.3 & 89.0 & 54.6
    \\
    FWA\cite{Li2018ExposingDV} & UADFV & \textbf{80.1} & 97.4 & 56.9
    \\
    VA-MLP\cite{Exploiting2019}  & Private & 66.4 & 70.2 & 55.0
    \\
    VA-LogReg\cite{Exploiting2019}  & Private & 78.0 & 54.0 & 55.1
    \\
    Multi-task\cite{nguyen2019multitask}  & FF & 76.3 & 65.8 & 54.3
    \\
    Xception\cite{rssler2019faceforensics}  & FF++ & 99.7 & \textbf{80.4} & 48.2
    \\
    Capsule\cite{nguyen2019use}  & FF++ & 96.6 & 61.3 & 57.5
    \\
    Xception\cite{Celeb_DF_cvpr20}  & UADFV & - & 96.8 & 52.2
    \\
    Xception+Reg.\cite{dang2019detection}  & UADFV & - & 98.4 & 57.1
    \\
    \midrule
    Xception+Tri.(ours)  & FF++ & \underline{99.9} & 74.3 & \textbf{61.7}
    \\
    Xception+Tri.(ours)  & UADFV & 61.3 & \underline{99.9} & \textbf{60.0}
    \\
    Xception+Tri.(ours)  & Celeb-DF & 60.2 & 88.9 & \underline{99.9}
    \\
    \bottomrule
    \end{tabular}
\end{table}
For FF++, we achieved a $0.2\%$ improvement comparing to the Xception \cite{rssler2019faceforensics} when testing the same dataset, reaching $99.9\%$. For the cross-dataset setting, while our method is weaker than Xception when testing UADFV, it boosts $13.5\%$ on the Celeb-DF which is more representative than the UADFV in terms of quality, quantity and diversity.

Regarding the detectors trained with UADFV, one strong competitor is FWA \cite{Li2018ExposingDV}, which is merely $2.5\%$ and $3.1\%$ less than our method when testing the UADFV and Celeb-DF datasets, respectively. However, FWA achieves an AUC $18.8\%$ higher than our method on testing the FF++ dataset. It is worth mentioning that FWA is based on the observations that previous face forgery algorithms could only generate images with limited resolutions, which requires to warp the fake face to match the original face. In other words, these kinds of methods \cite{Li2018ExposingDV,Yang2019ExposingDF} detect the forgery faces based on the imperfections of the manipulation algorithms, which can easily be obsoleted. This disadvantage becomes significant on the Celeb-DF. In this regard, our method has a better generalization ability. 

We also observe that when training on UADFV, method \cite{dang2019detection} ranks the 2nd and the 3rd place when testing on UADFV and Celeb-DF, respectively. We also show outstanding performance of our method on the Celeb-DF which is a relatively new, larger dataset in higher resolution with refined videos. Note that, it provides noticeably different information for the network to learn. When testing the other two datasets, the loss of such information may cause the variation on the performance. Our method outperforms state-of-the-art techniques, in terms of the intra-dataset setting (underlined numbers), and generates best results on testing the more challenging Celeb-DF in cross-dataset setting (bold numbers).

\section{Conclusion}
We have introduced a deep learning approach for the detection of face manipulation. It first learns the contrastive features which are then taken as input to a simple binary classification network. Results show that our method is effective, and outperforms state-of-the-art techniques in most cases. Since there are still noticeable space for improving the cross-dataset detection performance, we would like to design more effective face manipulation detection techniques with high generalization capability in the future.

%
%
%
%

\end{document}